\documentclass[letterpaper]{article} 
\usepackage{aaai24}  
\usepackage{times}  
\usepackage{helvet}  
\usepackage{courier}  
\usepackage[hyphens]{url}  
\usepackage{graphicx} 
\usepackage{natbib}  
\usepackage{caption} 
\usepackage{algorithm}
\usepackage{algorithmic}
\usepackage{comment}

\usepackage{epstopdf}
\usepackage{multirow}
\usepackage{amsmath}
\usepackage{mathtools}
\usepackage{cite}
\usepackage{multicol}
\usepackage{graphicx}
\usepackage{latexsym}
\usepackage[justification=centering]{caption}
\usepackage{amsthm}
\usepackage{relsize}
\usepackage{color}
\usepackage{amssymb}
\usepackage{eucal}
\usepackage{url}
\usepackage{comment}
\usepackage{booktabs,subcaption,amsfonts,dcolumn}
\usepackage{lipsum}

\usepackage{pifont}

\usepackage{natbib}  

\usepackage[textwidth=3cm,textsize=footnotesize]{todonotes}
\newcommand{\eric}[1]{\todo[color=red!20]{{\bf EM:} #1}}
\newcommand{\erici}[1]{\todo[inline,color=red!20]{{\bf EM:} #1}}
\newcommand{\tcr}[1]{\textcolor{red}{#1}}

\usepackage{cleveref}

\usepackage{enumerate}
\usepackage{csquotes}
\usepackage{calrsfs}
\DeclareMathAlphabet{\pazocal}{OMS}{zplm}{m}{n}

\usepackage{bm}

\usepackage{todonotes}
\definecolor{PalePurp}{rgb}{0.66,0.57,0.66}

\def\argmax{\mathop\text{argmax}}

\usepackage{mathrsfs}

\usepackage{newfloat}
\usepackage{listings}
\DeclareCaptionStyle{ruled}{labelfont=normalfont,labelsep=colon,strut=off} 
\floatstyle{ruled}
\newfloat{listing}{tb}{lst}{}
\floatname{listing}{Listing}
\title{Deep Reinforcement Learning
Algorithms for Hybrid V2X Communication:\\ A Benchmarking Study}
\author{
    Fouzi Boukhalfa\equalcontrib$^{\rm 1}$,
    Reda Alami\equalcontrib$^{\rm 1}$,
    Mastane Achab$^{\rm 1}$, 
    Eric Moulines$^{\rm 2,3}$, 
    Mehdi Bennis$^{\rm 4}$
}
\affiliations{
    \textsuperscript{\rm 1}Technology Innovation Institute, 9639 Masdar City, Abu Dhabi, United Arab Emirates\\
    \textsuperscript{\rm 2}Ecole Polytechnique, Palaiseau, France\\
    \textsuperscript{\rm 3}Mohamed bin Zayed University of Artificial Intelligence, Abu Dhabi, United Arab Emirates\\
    \textsuperscript{\rm 4}University of Oulu, Finland\\
    fouzi.boukhalfa@tii.ae, reda.alami@tii.ae

}

\usepackage{bibentry}

\begin{document}

\maketitle

\begin{abstract}
In today's era, autonomous vehicles demand a safety level on par with aircraft. Taking a cue from the aerospace industry, which relies on redundancy to achieve high reliability, the automotive sector can also leverage this concept by building redundancy in V2X (Vehicle-to-Everything) technologies. Given the current lack of reliable V2X technologies, this idea is particularly promising. By deploying multiple RATs (Radio Access Technologies) in parallel, the ongoing debate over the standard technology for future vehicles can be put to rest. However, coordinating multiple communication technologies is a complex task due to dynamic, time-varying channels and varying traffic conditions. This paper addresses the vertical handover problem in V2X using Deep Reinforcement Learning (DRL) algorithms. The goal is to assist vehicles in selecting the most appropriate V2X technology (DSRC/V-VLC) in a serpentine environment. The results show that the benchmarked algorithms outperform the current state-of-the-art approaches in terms of redundancy and usage rate of V-VLC headlights. This result is a significant reduction in communication costs while maintaining a high level of reliability. These results provide strong evidence for integrating advanced DRL decision mechanisms into the architecture as a promising approach to solving the vertical handover problem in V2X.

\end{abstract}

\section{Introduction}

In recent years, V2X (Vehicle-to-Everything) communication has relied mainly on two technologies: a) dedicated short-range communication (DSRC)-based vehicular networks ~\cite{DSRC} and b) cellular vehicular networks (C-V2X) ~\cite{CellularBasedVN}. The integration of multiple technologies has enabled heterogeneous vehicular communication, which has attracted much attention and witnessed notable advancements. In addition to standard V2X communications, there is third emerging technology that is currently being standardized: Vehicular Visible Light Communications (V-VLC)~\cite{memedi2020vehicular}.
V2X has stringent quality of service (QoS) requirements, including very low end-to-end latency (less than 1 ms) and high transmission rates for large-scale data exchange (up to 1000 Mb/s). In addition, certain use cases require ultra-reliable connectivity with reliability rates of up to 99.999999\%, which cannot be consistently achieved under all traffic and weather conditions with a single technology. Currently, connected and autonomous communications systems rely primarily on a single communications link. While the United States and China have adopted C-V2X as the primary technology for connected and autonomous vehicles, Europe and other regions continue to debate the choice of technology.
In the existing literature, researchers have identified several research topics within the C-V2X stack that can benefit from the application of machine learning techniques (ML) to improve performance. These topics include resource allocation at the MAC level for the fully distributed mode of C-V2X, energy efficiency with particular focus on base stations (BSs) as the components with the highest energy consumption \cite{zhang20196g}, beam prediction for mmWave communications, and handover techniques.
Handover procedures in C-V2X include two types: horizontal and vertical handover. Horizontal handover occurs when a vehicle changes the serving base station (BS) due to its mobility. This handover is performed over the $X_2$ and $X_n$ interfaces for LTE-V2X and 5G NR V2X, respectively. Vertical handover (VHO), on the other hand, switches between different V2X technologies within the same node, which is also referred to as hybrid communication. This work focuses primarily on vertical handover, which is the main subject of investigation.
%
In this paper we propose a comprehensive benchmark of Deep Reinforcement Learning adapted for the unique characteristics of hybrid V2X technologies. This benchmark is crucial to establish an effective and reliable approach for solving the vertical handover problem in V2X communications. Deep RL is particularly suited for this problem given its ability to learn optimal policies in dynamic and complex environments, handling the trade-off between exploitation and exploration. Unlike traditional machine learning algorithms that rely on predefined features or supervised labels, Deep RL algorithms are able to continuously adapt their strategy based on the real-time reward signals from the environment, making them an ideal choice for the dynamic and uncertain V2X communication scenarios.

\subsection{Related Work}

Extensive research and applications of Dynamic Programming (DP), Model Predictive Control (MPC), and Machine Learning (ML) have been conducted in the area of vehicle-to-everything (V2X) communication systems. These approaches show great promise in optimizing communication performance and resource allocation in V2X scenarios.
DP, a well-established optimization method, has been effectively used to solve various V2X communication problems. By decomposing complex problems into smaller subproblems, the algorithms of DP enable efficient computation of optimal strategies. For example, DP has been used to optimize transmission power and resource allocation in V2X networks, taking into account important factors such as channel conditions, interference, and quality of service requirements. The application of DP-based approaches provides valuable insights into optimal decision strategies within V2X communication systems \cite{DPV2X}.
Model Predictive Control (MPC) is another popular technique that has been used to improve V2X communication performance. MPC uses predictive models of system dynamics to optimize control actions over a finite horizon, subject to constraints. In the context of V2X, MPC has been used to optimize transmit power, antenna beamforming, and resource allocation to maximize throughput, minimize latency, and improve reliability. MPC's ability to account for system dynamics and incorporate real-time measurements makes it well-suited for dynamic V2X environments \cite{MPCV2X1,MPCV2X2,MPCV2X3,MPCV2X4}.
Machine learning techniques are playing an increasingly important role in optimizing V2X communications. Among these important techniques are (self)-supervised learning, and deep reinforcement learning (DRL). These machine learning algorithms excel at seeking communication policies from vast amounts of data, enabling autonomous decision making in V2X systems. The use of machine learning allows V2X systems to dynamically refine their communication strategies. This flexibility relies on the ability of these systems to evaluate the observed environment and take into account historical data. In particular, supervised learning methods have proven to be particularly effective in predicting channel conditions. DRL played a critical role in optimizing resource allocation and dynamically adjusting communication parameters driven by real-time measurements.
%
%
%
%
Reinforcement learning emerges as a potent methodology for sequential decision-making under uncertainty. It involves an agent interacting with an environment, learning from feedback in the form of rewards or penalties, and adapting its behavior to maximize cumulative rewards over time. This learning process enables the agent to make optimal decisions in complex, dynamic, and uncertain scenarios.
In such environments, the agent needs to handle a continuous stream of sensory inputs and make decisions based on incomplete or noisy information. Reinforcement learning provides a framework to model and address these challenges. By employing a trial-and-error learning approach, the agent explores different actions and learns from their outcomes, gradually refining its decision-making strategy.

\subsection{Contributions}
This paper presents a comprehensive benchmark of different Deep Reinforcement Learning (DRL) algorithms such as Proximal Policy Optimization (PPO \cite{PPO}), Trust Region Policy Optimization (TRPO \cite{TRPO}), Rainbow Deep Q Networks (Rainbow DQN \cite{hessel2018rainbow}) and Soft Actor-Critic (SAC \cite{SAC}) used in a V2X simulator to manage the handover process. The study models the handover as a Markov Decision Process (MDP) and using the capabilities of these complex DRL algorithms in a simulated environment. A unique feature of this paper is the introduction of a generalization benchmark to compare the adaptability and performance of each DRL algorithm in a variety of network conditions and configurations. The focus is to maximize the transmission reliability and resource efficiency, which are critical parameters for the handover decision. For each algorithm, we provide a deep analysis and a comparison in term of the performance, stability, complexity and hysteresis effect.


\section{Problem formulation}



The advent of V2X communications is paving the way for a safer, more efficient, and smarter transportation system. 
Central to this evolution is the ability of vehicles to communicate effectively and efficiently with each other and with the surrounding infrastructure. In the vehicular network, it is possible to approach the full capacity of the channel when the density of vehicles increases. When this occurs, some vehicles will be unable to communicate with the rest of the network. Furthermore, in the countryside, some area may be not covered by the C-V2X. In these situations, combining radio frequency (RF) and V-VLC will be a suitable solution. When a degradation of the QoS for radio technology or another is detected, a VHO procedure should be initiated. Powered this solution by DRL agent allows the maintenance of high reliability that satisfy advanced autonomous vehicle applications. However, the task  is not straightforward. The different communication technologies differ in terms of their cost and success probability, which depend on the dynamic conditions of the vehicle environment. 
In addition, the agent has the option to use redundant communication technologies to increase the success probability. This results in a complex decision problem that requires careful consideration of communication success and cost.
Our problem can be modeled as a Markov decision process (MDP) in the RL framework characterized by a state space, an action space, probability transitions and a reward function. The action space consists of the possible choice of communication technologies. The reward function balances the probability of communication success and the cost of each action, with the goal of maximizing the former and minimizing the latter.
In this way, we seek a strategy that optimizes the long-term return of the RL agent's actions, taking into account the dynamic nature of vehicle communication. By learning this strategy, the agent can make informed decisions on the best communication technology to use on the fly, maximizing the efficiency and reliability of V2X communications.




\section{Benchmark Proposal: Optimal control using Deep Reinforcement Learning}

In this section, we first describe the reinforcement learning algorithms selected for the benchmark and then the protocol of the benchmark.

\subsection{Value based and distributional strategies}
The Rainbow DQN (Rainbow Deep Q-Network \cite{hessel2018rainbow}) algorithm is an advanced variant of the original DQN algorithm that combines several state-of-the-art techniques to enhance the training process and improve the performance of reinforcement learning agents. It incorporates a set of key components, including prioritized experience replay, dueling network architectures, multi-step learning, distributional reinforcement learning, and NoisyNet exploration.

In \textbf{prioritized experience replay}, the algorithm assigns priorities to each experience tuple based on its estimated TD-error, which is calculated using the Bellman equation. This prioritization allows the agent to replay experiences that are deemed more informative or challenging, leading to more effective learning.

The architecture of \textbf{duelling network} decouples the estimation of the value and advantage functions and allows the agent to better understand the value of each action in different states. The value function represents the expected return of a given state, while the advantage function quantifies the benefits of each action. The final action value estimates are determined by combining the value and advantage functions.

\textbf{Multi-step learning} extends the original DQN by considering a sequence of actions instead of just single steps. This helps the agent to capture longer-term dependencies and improves learning efficiency.

\textbf{Distributional reinforcement learning} introduces a distributional perspective in which the agent does not estimate the expected return, but learns the distribution of possible returns for each action. This approach provides a more complete representation of uncertainty and variability in the environment and allows for more robust decision making.

\textbf{NoisyNet} exploration introduces stochasticity into the weights of the network, allowing the agent to explore a wider range of actions during training. Adding random noise to the parameters of the network allows the agent to discover new and potentially better strategies.

By combining these techniques, Rainbow DQN achieves improved exploration, faster convergence, and enhanced generalization capabilities that enable reinforcement learning agents to more effectively cope with complex and high-dimensional environments.

\subsection{Policy gradient based strategies}
Policy gradient is a class of RL algorithms based on the maximization of the following objective function:
\begin{equation}
\label{eq:Jtheta}
J(\theta)=\mathbb{E}\left[\sum_{t=0}^T \gamma^t r\left(s_t, a_t\right) \right],
\end{equation}
where $\theta$ parameterizes (via a deep neural network) the agent policy.
We describe below a few policy gradient methods: roughly speaking, they all maximize the quantity $J(\theta)$ by gradient ascent.

\paragraph{Proximal Policy Optimization (PPO)}
Proximal Policy Optimization (PPO) \cite{PPO} is an algorithm used in reinforcement learning to optimize policies by balancing exploration and exploitation. It uses a surrogate objective function and a trust region constraint to update policy parameters. The objective of PPO is to maximize a surrogate variant of the expected cumulative reward $J(\theta)$. The surrogate objective function $L(\theta)$ is an approximation of the expected performance improvement:
\begin{equation}
\label{eq:ppo_clip}
\scalebox{0.84}{$
 L(\theta)=  
 \mathbb{E}\left[\min \left(\frac{\pi_\theta(a \mid s)}{\pi_{\theta_{\text {old }}}(a \mid s)} A(s, a),  \operatorname{clip}\left(\frac{\pi_\theta(a \mid s)}{\pi_{\theta_{\text {old }}}(a \mid s)}, 1-\epsilon,  1+\epsilon\right) A(s, a)\right)\right]  .
 $}
\end{equation}

Here, $\pi_\theta(a \mid s)$ and $\pi_{\theta_{\text {old }}}(a \mid s)$ represent the probabilities of performing an action $a$ in state $s$ under the current and previous policies, respectively. $A(s, a)$ denotes the advantage function, which measures the advantage of performing action $a$ in state $s$ compared to the average action value. The clip function limits policy updates within the range $(1-\epsilon, 1+\epsilon)$ to ensure stable updates. PPO performs multiple iterations of stochastic gradient ascent to maximize the surrogate objective function while staying within the trust region. By utilizing the surrogate objective function and trust region constraint, PPO achieves stable and efficient policy updates, striking a balance between exploration and exploitation in reinforcement learning tasks.

\paragraph{Trust Region Policy Optimization (TRPO)}
Trust Region Policy Optimization (TRPO) \cite{TRPO} is a reinforcement learning algorithm that aims to optimize policies while maintaining a trust region to ensure stable policy updates. It uses a surrogate objective function and a constraint on maximum policy divergence.
The objective of TRPO is to maximize this surrogate version of $J(\theta)$. The surrogate objective function, $L(\theta)$, is defined as:
\begin{equation}
\label{eq:trpo}
L(\theta)=\mathbb{E}\left[\frac{\pi_\theta(a \mid s)}{\pi_{\theta_{\text {old }}}(a \mid s)} A(s, a)\right].
\end{equation}
TRPO introduces a constraint on policy updates to guarantee a trust region. The constraint ensures that the updated policy does not deviate too far from the previous policy. The constraint is expressed as follows:
\begin{equation}
\label{eq:trpo_cons}
\mathbb{E}_{s \sim \rho_{\theta_{\text {old }}}}\left[\mathrm{KL}\left(\pi_{\theta_{\text {old }}}( \cdot \mid s) \| \pi_\theta( \cdot \mid s)\right)\right] \leq \delta.
\end{equation}
Here, $\mathrm{KL}$ represents the Kullback-Leibler divergence, $\rho_{\theta_{\text {old }}}$ is the state distribution under the previous policy, and $\delta$ is the maximum allowed policy divergence. TRPO solves the optimization problem by iteratively updating the policy within the trust region. It utilizes a line search to find the step size that maximizes the objective function while satisfying the trust region constraint. By maintaining a trust region and performing constrained optimization, TRPO ensures stable policy updates and achieves better sample efficiency compared to other methods. It strikes a balance between exploration and exploitation, leading to more reliable and efficient policy optimization.

\paragraph{Soft Actor Critic (SAC)}

The Soft Actor-Critic (SAC) algorithm \cite{SAC} is a state-of-the-art reinforcement learning algorithm that is mainly devoted to learning optimal policies in continuous action spaces (though it can still be used in our discrete context). SAC combines the advantages of policy optimization and value-based methods while addressing the exploration-exploitation trade-off. It utilizes a maximum entropy framework to encourage exploration and maintain a distributional representation of the policy. The objective of SAC is to maximize the expected cumulative reward, denoted as $J(\theta)$. The algorithm introduces an entropy regularization term to the objective function, promoting exploration:
\begin{equation}
\label{eq:sac_entropy}
\widetilde{J}(\theta)=\mathbb{E}\left[\sum_{t=0}^T \gamma^t\left(r\left(s_t, a_t\right)+\alpha \mathcal{H}\left(\pi\left(\cdot \mid s_t\right)\right)\right)\right].
\end{equation}
Here, $r\left(s_t, a_t\right)$ represents the immediate reward at time step $t, \gamma$ is the discount factor, $\pi\left(\cdot \mid s_t\right)$ denotes the policy distribution over actions given state $s_t, \mathcal{H}\left(\pi\left(\cdot \mid s_t\right)\right)$ represents the entropy of the policy distribution, and $\alpha$ is a temperature parameter that balances the entropy regularization. SAC employs twin Q-networks to estimate the action-value function. The two Q-networks (with respective set of parameters $\theta_1$ and $\theta_2$) help to mitigate the overestimation bias commonly found in value-based methods. The value function loss is defined as the mean squared Bellman error:

\begin{equation}
\mathcal{L}_V(\theta)=\mathbb{E}\left[\frac{1}{2}\left(Q_{\theta_1}\left(s_t, a_t\right)-y_t\right)^2\right]   
\end{equation}

where $Q_{\theta_1}\left(s_t, a_t\right)$ is the estimate of the action-value function, and $y_t=r\left(s_t, a_t\right)+\gamma(1-$ $\left.d_t\right) V_{\theta_2}\left(s_{t+1}\right)$ represents the TD target, incorporating the next state value estimate $V_{\theta_2}\left(s_{t+1}\right)$ and a discount factor $d_t$ that masks out terminal states.
To update the policy, SAC employs the reparameterization trick, which enables backpropagation through stochastic actions. The policy loss is defined as the negated expected state-action value plus an entropy regularization term:
\begin{equation}
\mathcal{L}_\pi(\theta)=\mathbb{E}\left[\alpha \mathcal{H}\left(\pi\left(\cdot \mid s_t\right)\right)-Q_{\theta_1}\left(s_t, \epsilon_t\right)\right]    
\end{equation}

where $\epsilon_t$ represents some noise sampled from the policy distribution.
SAC also utilizes a target network for stable value function updates. It employs a soft update mechanism to slowly update the target networks towards the main networks.
By jointly optimizing the policy and value functions while incorporating the entropy regularization, SAC achieves effective exploration and learns optimal policies in continuous action spaces. The algorithm has demonstrated impressive performance in a wide range of challenging reinforcement learning tasks.

\subsection{Benchmark}

In the benchmark of the vehicular communication problem using the described MDP model, four DRL algorithms are evaluated: PPO, TRPO, Rainbow DQN, and SAC. These algorithms are applied to the serpentine simulation environment described in the next section. The results provide insights into the effectiveness and suitability of each algorithm for addressing the vehicular communication problem.


\section{Simulation results}

In this section, we first describe the serpentine scenario using V2X technology. Then, we describe the training process and analyse the performances of each deep reinforcement learning algorithm. Finally, we discuss the robustness and complexity of each algorithm.

\subsection{Serpentine scenario using V2X technologies}
\label{subsec:serpentine}
We used the two scenarios that was defined in \cite{schettler2020train}. These scenarios involve two vehicles following each other on a very curvy road and communicating via V2X technologies (VLC taillight, VLC headlight, and DSRC). The nature of these environments, generated via SUMO~\cite{behrisch2011sumo}, make both technologies in a challenging situations. Table~\ref{tab:Simulation} summarizes the simulation parameters used in our scenarios.

\begin{table}[!htbp]
\begin{center}
\caption{\textsl{Simulation parameters}.}
\renewcommand{\arraystretch}{0.4}
\begin{tabular}{cc}
    \hline
  \textbf{Parameter}&\textbf{Value} \\ 
      \hline
 &  \\
												
\textbf{\textit{Packet Byte length}}& $1024 \ \text{byte}$ \\	
 &  \\												
\textbf{\textit{Beaconing Frequency}}& 10Hz \\	
                  &  \\
\textbf{\textit{
Transmission power (DSRC) }}& 20mW \\
&  \\
 \textbf{\textit{Bitrate (DSRC) / (VLC)}}& 6Mbps / 1Mbps \\
 &  \\
 \textbf{\textit{Vehicles speed}}& 30 - 40km/h \\
  &  \\
 \textbf{\textit{Simulation time}}& 400s \\

 \hline
 \label{tab:Simulation}
 \end{tabular}
 \end{center}
\end{table}

The follower is the one who owns the agent. For each message transmitted to the leader, it has to select the optimal communication link based on the observation (see ~\Cref{fig:angles}). The first scenario was used to train and test the different algorithms: Serpentine with few hairpin curves, while the second scenario with many hairpin curves was dedicated to robustness testing.

\subsection{Reinforcement learning setting}
In this section, we start by describing the markovian decision process used in the benchmark and its limitations.
\paragraph{Model description}

\begin{figure}[ht]
\includegraphics[width=\columnwidth, ,height=4cm]
{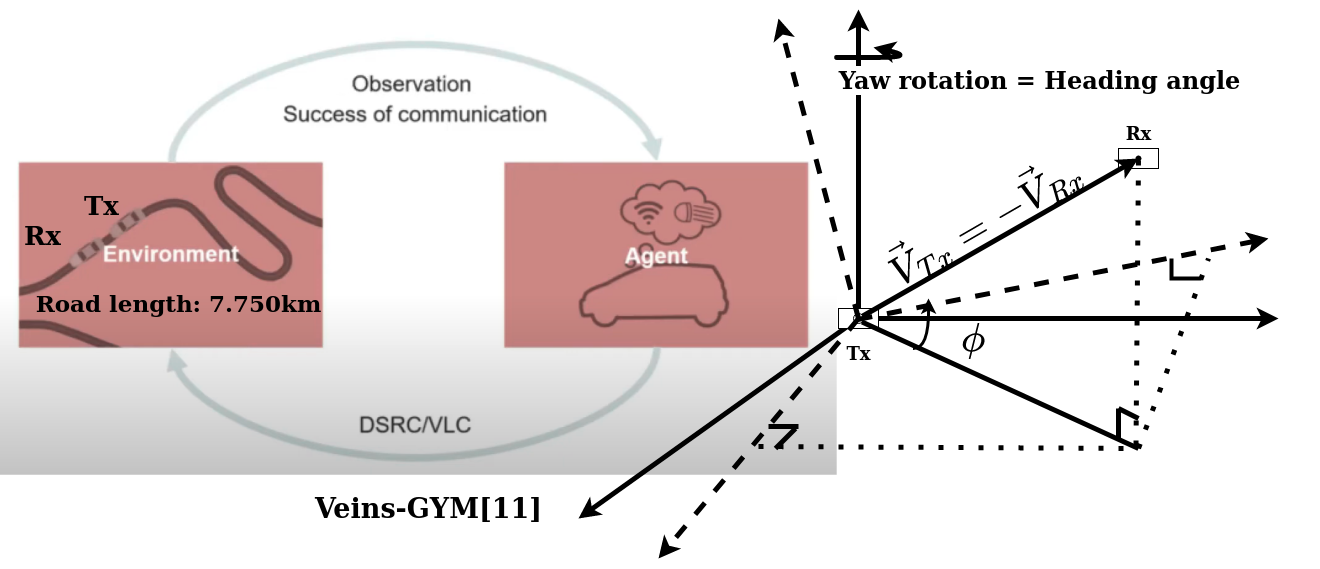}
\caption{Environment observed by the agent.}
\label{fig:angles}
\end{figure}

We consider a reinforcement learning approach to our vehicle communication problem through a Markov decision process (MDP) model characterized by:
\begin{itemize}
 \item State space $\mathcal{S} \subseteq \mathbb{R}^4$ (\Cref{fig:angles}): where the state $s \in \mathcal{S}$ is described as: $s = \Big[X = \text{clip}\Big(
 \frac{x_{\Vec{V}_{Tx}}}{R}, -1, 1 \Big), Y = \text{clip}\Big(
 \frac{y_{\Vec{V}_{Tx}}}{R}, -1, 1 \Big), \cos(\phi) = \frac{x_{\Vec{V}_{Rx}}}{\| \Vec{V}_{Rx} \|} ,          \sin(\phi) = \frac{y_{\Vec{V}_{Rx}}}{\| \Vec{V}_{Rx} \|}\Big]$,
 


where $R$ is the maximum transmission range of DSRC (in practice, $R=10^{3}m$), and $R_x$, $T_x$, $\Vec{V}_{Rx}$, $\Vec{V}_{Tx}$ refer respectively to the receiver, the transmitter and their positional vectors as shown in ~\Cref{fig:angles}.

 \item Action space $\mathcal{A}=\{ a_1, a_2, \dots, a_8 \}$ where the 8 possible actions correspond to: No transmission ($a_1$), DSRC ($a_2$), VLC Headlight ($a_3$), DSRC \& Headlight ($a_4$), VLC Taillight ($a_5$), DSRC \& Taillight ($a_6$), Taillight \& Headlight ($a_7$), all available technologies ($a_8$).
 \item Probability transition $P(s'|s, a)$ between states generated by the environment. In our case, the chosen communication channel does not affect the trajectories of the cars, thus the transition kernel simplifies to $P(s'|s)$.
 \item Average reward function $r(s, a) =p(s, a) - C(a)$ balancing the communication success probability $p(s, a)$ and the communication cost $C(a)$.
\end{itemize}

If the agent performs an action $a\in\mathcal{A}$ while the environment is in state $s\in\mathcal{S}$, then the next state $s^\prime \sim P(\cdot|s,a)$ is sampled from the distribution $P(\cdot|s,a)$ and the expected immediate reward is $r(s,a)$.

\paragraph*{\textbf{Environment limitations}}
Describing the state in the V2X communication problem as $s=(X, Y, \cos\phi, \sin\phi)$ has some limitations. While this state representation captures the position and angle of the agent (vehicle), it may not fully capture all the relevant information for effective V2X communication. Here are some limitations of this state representation:
\begin{itemize}
 \item Limited information about the environment: the state representation contains only the agent's position and angle. It does not provide explicit information about the presence or location of other vehicles, road conditions, traffic congestion, signal interference, or other environmental factors that may significantly affect V2X communications.
 \item Historical context: the state representation does not contain historical information about the agent's past states or actions. However, historical context can be critical to understanding the agent's trajectory, its interaction with other vehicles, and the effectiveness of previous communication attempts.
 \item Incomplete channel properties: The state representation does not take into account specific characteristics of the communication channel, such as signal strength, channel quality, interference level, or available bandwidth. These factors are critical to evaluating the feasibility and reliability of inter-vehicle communications.
\end{itemize}

\subsection{Training}

\begin{figure}[ht]
\includegraphics[height=5 cm ,width=\columnwidth]
{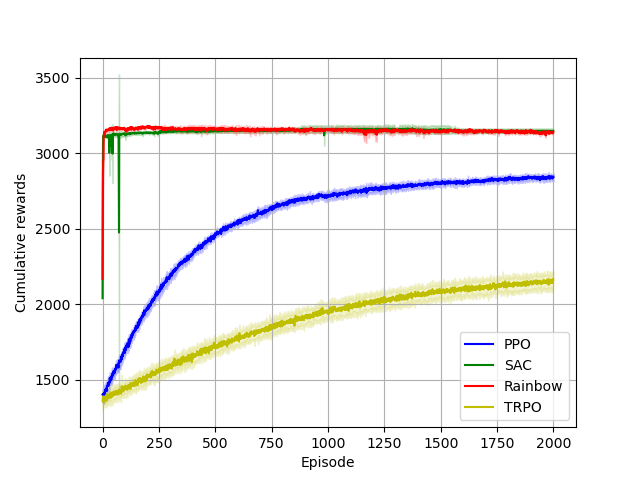}
\caption{Learning curves on the V2X
simulated environment with 95\% confidence intervals for five runs (seed)}
\label{Benchmark_training}
\end{figure}

We train the four deep RL agents (PPO, TRPO, Rainbow DQN, SAC) using scenario $1$ and we fine tune them with a grid search on the main hyperparameters (see ~\Cref{Grid_search}).
\Cref{Benchmark_training} displays the learning curve for each model trained in the simulation environment. In the following, we describe the exploration-exploitation tradeoff that explains the convergence of the learning curves displayed.

\begin{itemize}
    \item \textbf{PPO}: ensures convergence by balancing exploration and exploitation. It introduces randomness in action selection using a Gaussian distribution while leveraging the current policy to estimate rewards accurately.
    \item \textbf{TRPO}: ensures convergence by employing a trust region approach. It limits policy updates to a trust region to balance exploration and exploitation, achieving significant improvements while maintaining stability.
    \item \textbf{SAC}: uses an actor-critic framework, where the actor explores the environment and the critic provides value estimates for balancing exploration and exploitation. Entropy regularization encourages diverse actions, preventing premature convergence to a deterministic policy.
    \item \textbf{Rainbow DQN}: actively explores the environment through techniques like epsilon-greedy exploration and experience replay. It balances exploitation using double Q-learning, prioritized experience replay, and the dueling network architecture to optimize policies effectively in complex environments
\end{itemize}

\begin{table}
    \centering
    \caption{Deep RL Benchmark Grid search.}
    \label{Grid_search}
    \begin{tabular}{ccc}
        \toprule
        Algorithm              & Hyper-parameter & Values 
        \\
        \midrule
        \multirow{3}{*}{RainbowDQN} & $\alpha$ (learning rate)        & \{$0.1$, $\mathbf{0.3}$\}        
        \\
                           & $\beta$ (Weight Decay)       & \{$0.5$, $\mathbf{0.7}$\}        
                           \\
                           & Atome size       & \{$\mathbf{25}$, $100$\}
         
                           \\
                           & Prior epsilon       & \{$1e^{-5}$, $\mathbf{1e^{-4}}$\}        
                           \\
        \midrule
        \multirow{2}{*}{TRPO} & $\delta$  (TR Upper Bound)      & \{$\mathbf{0.005}$, $0.01$\}        
        \\
                           & Depth of neural nets       & \{$32$, $\mathbf{64}$\}        
                           \\
                           & line search max iter        & \{$10$, $\mathbf{20}$\}        
                           \\
        \midrule
        \multirow{2}{*}{SAC} & $\tau$ (soft update coefficient)       & \{$\textbf{0.005}$, $1e^{-2}$\}        
        \\
                           & Learning rate        & \{$3e^{-4}$, $\mathbf{5e^{-4}}$\}        
                           \\
                             & Batch size       & \{$\mathbf{64}$, $128$, $256$\}        
                           \\
        \midrule
        \multirow{2}{*}{PPO} & Learning rate Actor       & \{$\mathbf{1e^{-5}}$, $1e^{-2}$\}         
        \\
                           & Learning rate Critic        & \{$\mathbf{1e^{-3}}$, $1e^{-2}$\}        
                           \\
                             & $\epsilon$ (Epsilon clip)        & \{$\mathbf{0.2}$, $0.3$\}        
                           \\
                       
        \bottomrule
    \end{tabular}
\end{table}

\begin{table}[t]
\scriptsize
\caption{Learned strategies (PPO, SAC, Rainbow DQN, TRPO best model).}
\renewcommand{\arraystretch}{0.4}
\label{tab:tab_actions}

\begin{tabular}{lllccc}
    \hline
  \textbf{V2X technologies}&\textbf{PPO }&\textbf{Rainbow DQN} &\textbf{TRPO} & \textbf{SAC}\\ 
      \hline
 &  \\			
 No transmission & 0 & 0 & 0.02 & 0.02 \\	
 &  \\							
 DSRC& $7\%$ & $8.27\%$ & $5.14\%$ & $8.29\%$\\	
 &  \\
 VLC Headlight & $91\%$ & $90.77\%$ & $94.82\%$ & $91.67\%$\\	
 &  \\
 DSRC \& Headligth& $1.5\%$ & $0\%$ & $0\%$ & $0\%$ \\	
 &  \\
 VLC Tailligth& $0\%$ & $0\%$ & $0\%$ & $0\%$\\	
 &  \\
 DSRC \& Tailligth & $0\%$ & $0.44\%$ & $0\%$ & $0\%$\\	
 &  \\
 Tailligth \& Headligth& $0\%$ & $0.38\%$ & $0\%$ & $0\%$ \\	
 &  \\
 All available technologies& $0\%$ & $0.11\%$ & $0\%$ & $0\%$ \\	
 &  \\
 \hline
\end{tabular}

\end{table}

\begin{table}[t]
\scriptsize
\begin{center}
\caption{Deep RL performances on scenario $1$. SoA means the multi-Q regressor algorithm in \cite{schettler2020train}.}
\renewcommand{\arraystretch}{0.4}
 \label{tab:tab1}
\begin{tabular}{lccccc}
    \hline
  \textbf{Metrics}&\textbf{PPO}&\textbf{SoA}&\textbf{SAC}&\textbf{Rainbow}&\textbf{TRPO} \\ 
      \hline
 &  \\
			
 Reliability
 & $98.49\%$ & $\textbf{99.5}\%$  & 98.52\% &  $97.57\%$ &  $95.51\%$  \\							
 VLC utilization rate &  $93.68\% $& 58\% & $92.50\% $  & $91.24\%$ & $\textbf{95.35}\%$ \\
No Redundancy
 & $99.97\%$ & $99.4\%$ 	
 & $\textbf{100}\% $ & $99.14\%$ & $\textbf{100}\%$ \\
Taillight rate
&  $0\% $&  $0\% $ & $0\% $  & $0.85\%$ & $0\% $ \\
Number of switch
&  $8 $&  $/ $ & $ 27$  & $29$ & $15 $ \\ 
 \hline
 
\end{tabular}
\end{center}
\end{table}

\begin{table}[t]
\scriptsize
\begin{center}
\caption{Performances on scenario $2$.}
\renewcommand{\arraystretch}{0.4}
 \label{tab:tab_perf}
\begin{tabular}{lcccc}
    \hline
  \textbf{Reliability}&\textbf{PPO}&\textbf{SAC}&\textbf{RainBow DQN}&\textbf{TRPO} \\ 
      \hline
 &  \\
			
 1$^{\text{st}}$ best model
 & $84.84\%$ & $\textbf{94.78}\%$  & $71.84\%$ &  $78.81\%$ \\							
 2$^{\text{nd}}$ best model &  $97.57\%$ & $\textbf{97.71}\%$ & $95.62\%$  & $95.27\%$ \\
 
 \hline
\label{table:Generalisation}
\end{tabular}
\end{center}
\end{table}

\begin{table}[H]
\tiny
\begin{center}
\caption{Selected architecture for each DRL algorithm.}
\renewcommand{\arraystretch}{0.4}
 \label{tab:tab_perf_complexity}
\begin{tabular}{lcccc}
    \hline
  \textbf{Metrics}&\textbf{PPO}&\textbf{SAC}&\textbf{RainBow DQN}&\textbf{TRPO} \\ 
      \hline
 &  \\
Nbr. trainable parameters
&  $9545$  & $276512$  & $62689$ & $1225$ \\
Actor network structure
&  $4$,$64$,$64$,$8$ & $4$,$256$,$256$,$8$  & / & $4$,$64$,$8$ \\
Critic network structure
&  $4$,$64$,$64$,$1$ & $4$,$256$,$256$,$8$  & / & $4$,$64$,$1$ \\
Value network structure
&  / & /  & $4$,$128$,$128$,$25$ & / \\
Advantage network structure
&  / & /  & $4$,$128$,$128$,$200$ & / \\
 \hline
\label{table:Complexity}
\end{tabular}
\end{center}
\end{table}

\subsection{Performance analysis}

\begin{figure}[ht]
\includegraphics[height=5.5 cm ,width=\columnwidth]
{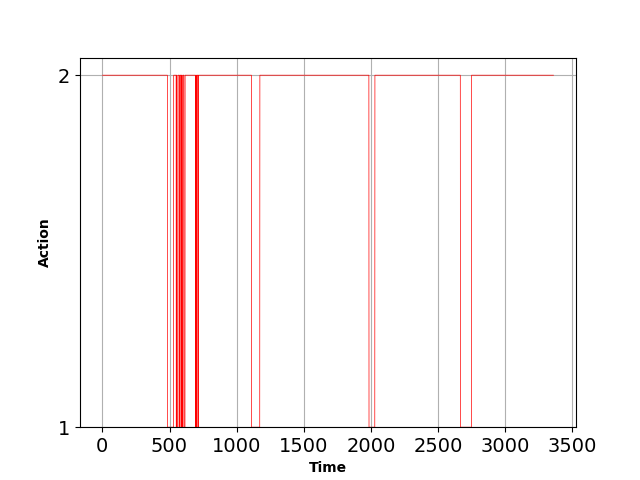}
\caption{SAC strategy: Action versus Time.}
\label{SAC_Action_Time}
\end{figure}

\begin{figure}[ht]
\includegraphics[height=5.5 cm ,width=\columnwidth]{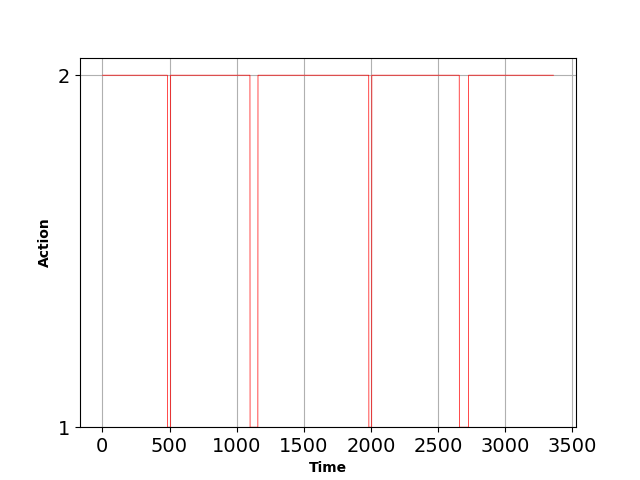}
\caption{TRPO strategy: Action versus Time.}
\label{TRPO_Action_Time}
\end{figure}

\begin{figure}[ht]
\includegraphics[height=5.5 cm ,width=\columnwidth]{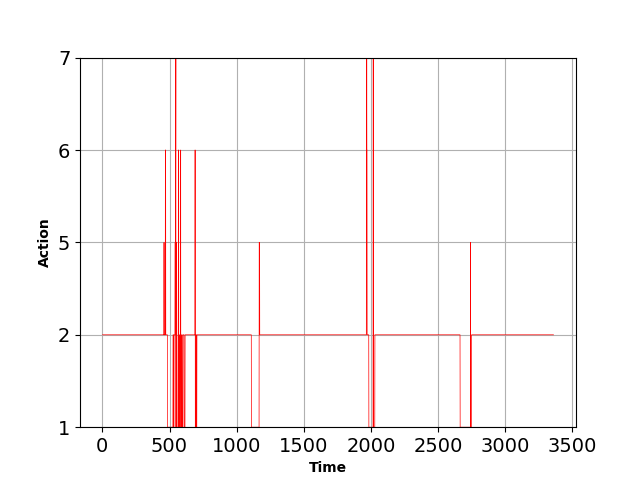}
\caption{Rainbow DQN strategy: Action versus Time.}
\label{RainbowDQN_Action_Time}
\end{figure}

\begin{figure}[ht]
\includegraphics[height=5.5 cm ,width=\columnwidth]{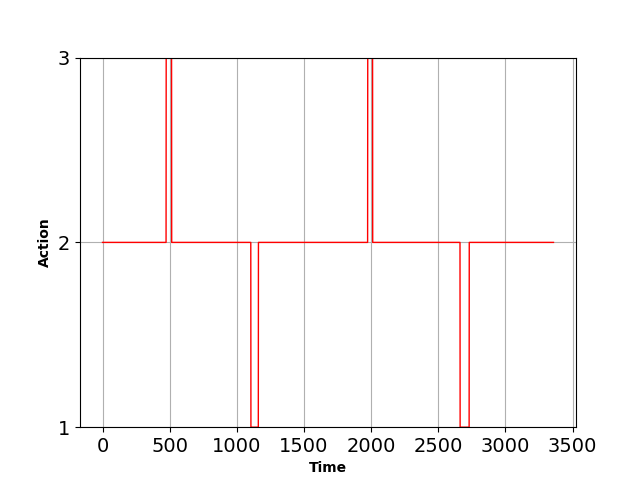}
\caption{PPO strategy: Action versus Time.}
\label{PPO_Action_Time}
\end{figure}

In this section we show the quality of each DRL agent. The agents must learn the following: maximize transmission reliability (expressed as Packet Delivery Ratio (PDR)), use no taillight (no vehicle behind), and minimize redundancy. Based on these observations, we establish four metrics to assess and quantify performance: reliability, VLC usage rate, no redundancy, taillight rate.
The baseline of this study was extracted from the paper of \cite{schettler2020train}. The authors adopted $5$ policies: DSRC/VLC only, hand-crafted heuristic formula and data-driven policies such as based on Convex-Hull formula and the deep RL Multi-Q-Regressor algorithm. \Cref{tab:tab1} summarizes these results. In Figures \ref{SAC_Action_Time},~\ref{TRPO_Action_Time},~\ref{RainbowDQN_Action_Time} and \ref{PPO_Action_Time}, we display the output action as function of the simulation time. Moreover, we provide in Table~\ref{tab:tab_actions} the percentage of each output action. We can conclude that the optimal action for all agents is VLC Headlight.
Furthermore, we see that PPO has the most stable learned policy, followed by the TRPO, SAC and Rainbow DQN. We can explain the policy stability through the entropy and policy constraints used in the learning process. The difference in policy stability between the PPO, TRPO, SAC, and Rainbow DQN when applied to vehicular communication can be related to their intrinsic characteristics in term of entropy and policy constraints. PPO and TRPO are methods with constrained policies (see Eqs. \ref{eq:ppo_clip} and \ref{eq:trpo_cons}), where the update step is limited to a trust region to ensure policy stability and prevent detrimental changes. This results in more consistent decisions regarding the choice of communication technology, thus leading to fewer switches. On the other hand, SAC employs an entropy regularization term (see Eq. \ref{eq:sac_entropy}), which encourages exploration of the action space. In the context of our vehicular communication scenario, this translates into more frequent shifts between communication technologies as the algorithm is encouraged to explore and exploit different options to find the optimal balance between reward and cost. Rainbow-DQN, while not directly utilizing entropy for exploration, combines several enhancements over standard DQN that may cause it to behave more variably in this context. The increased exploration encouraged by SAC and Rainbow-DQN may lead to higher reward in some settings, but in this case, it seems to cause more fluctuation in the choice of communication technology compared to PPO and TRPO. In Table \ref{tab:tab1}, we compute the number of switches for each agent. This metric quantify the stability described in the previous paragraph. Because even if the SAC gives better performance, the PPO outperform it in term of number of switch. In the design of vertical handover it is important to consider the sensitivity of the switching mechanism to not impact to much the End-To-End (E2E) delay. 



\begin{figure}[ht]
\includegraphics[width=\linewidth]{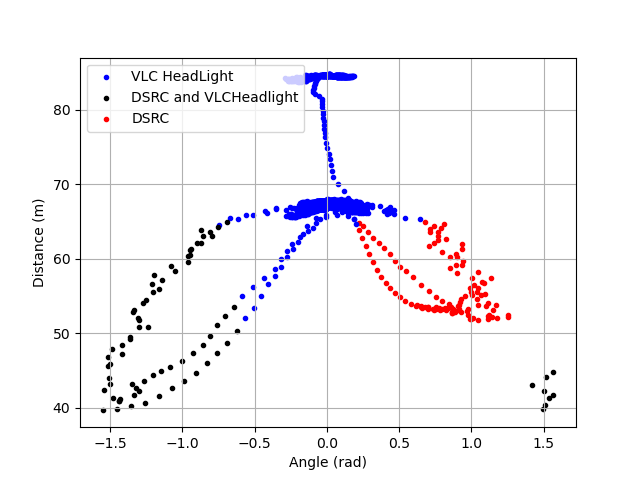}
\caption{PPO learned strategy.}
\label{State_Action_PPO}
\end{figure}

\begin{figure}[ht]
\includegraphics[width=\linewidth]{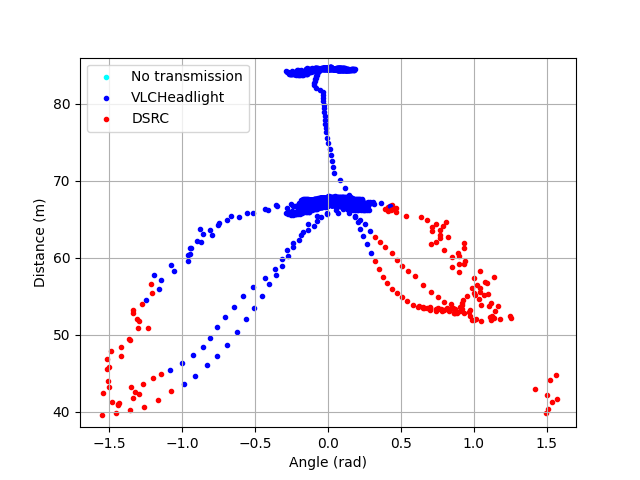}
\caption{TRPO learned strategy.}
\label{State_Action_TRPO}
\end{figure}

\begin{figure}[ht]
\includegraphics[width=\linewidth]{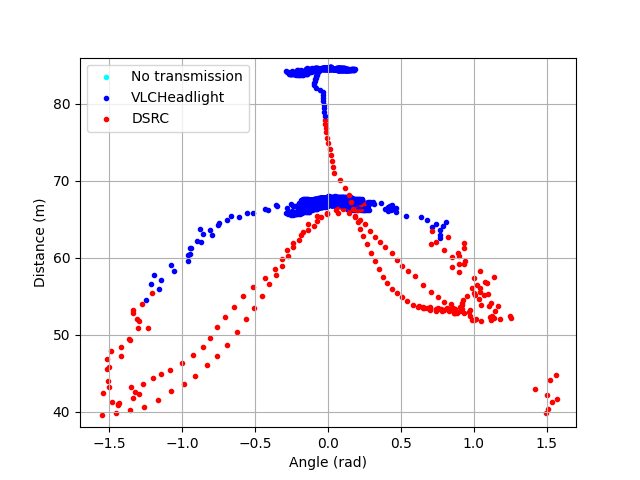}
\caption{SAC learned strategy.}
\label{State_Action_SAC}
\end{figure}

\begin{figure}[ht]
\includegraphics[width=\linewidth]{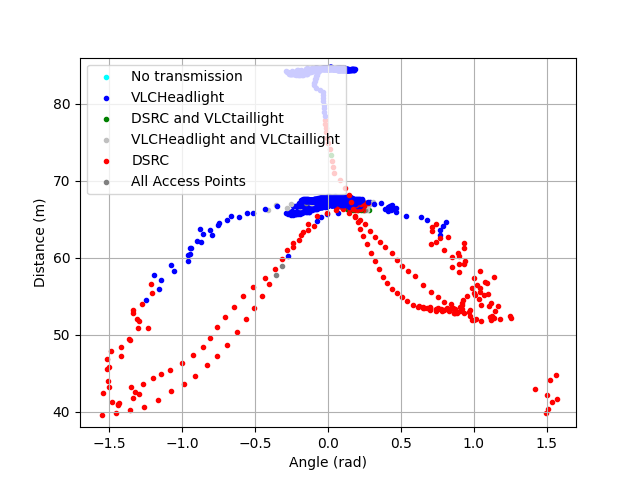}
\caption{Rainbow DQN learned strategy.}
\label{State_Action_Rainbow}
\end{figure}

Finally, from \cref{State_Action_PPO} to \cref{State_Action_Rainbow}, we plot the actions as a function of the states (distance \& angle). These figures can be used to illustrate the strategy chosen for each state. The simulation tool for V-VLC used in this section is based on\textbf{ VEINS-VLC}~\cite{VEINS-VLC}~\cite{VLC-platooning}. This simulator uses an empirical model. Since the data comes from a real headlight module, this radiation pattern includes the effect of high beam and low beam coming from real vehicular headlight modules. As we can observe, the agent chooses to use the VLC when the leader and follower are in Line Of Sight (LOS), since the conditions for its use are favorable and it's less costly (as defined in the reward design). Also, we can observe an asymmetric decision to use VLC respectively to the zero angle, this caused by asymmetric light distribution. Indeed, the manufactures design the low beam for this effect, the purpose is to not disturb the oncoming traffic from the opposite direction while properly illuminating the road in front of the vehicle. This will depend on the rule adopted in each country.

We can also see the boundaries of each decision. On these boundaries, we can see that there is some overlapping between actions. This creates a hysteresis effect (i.e. although some points represent almost the same state, the agent takes different decision), and this is the main cause of the ping-pong effect~\cite{4349970}, which makes the switch too sensitive and leads to unnecessary transitions. In telecommunications, two parameters are considered when designing VHO: hysteresis and Time To Trigger (TTT).  In other words, we should find the handover hysteresis margin that limits the ping-pong effect, this features should be taken into account in the reward design.

Table \ref{tab:tab_perf_complexity} describes the space complexity of the different algorithms that are used in our benchmark.
We selected these specific architectures by following the default structures from standard libraries, namely CleanRL \cite{huang2022cleanrl} and Acme \cite{hoffman2020acme}.
We point out that the two methods with the largest number of parameters, namely SAC and Rainbow-DQN, yield the best performance (see Figure \ref{Benchmark_training}). 
 


\subsection{Empirical sample complexity}


In deep reinforcement learning, 'sample complexity' refers to the minimal number of episodes necessary for an algorithm to converge. As illustrated in figure \ref{Benchmark_training}, SAC and Rainbow showcase notably lower sample complexities, needing fewer than 100 episodes, in contrast to TRPO and PPO. This superior performance is largely attributed to the robust exploration strategies of SAC and Rainbow DQN. Specifically, while TRPO and PPO are designed around trust region methods and policy gradient improvements respectively, SAC incorporates entropy into the reward, promoting more explorative policies. Similarly, Rainbow combines several advancements in DQN architecture, such as prioritized experience replay and dueling networks, enabling a richer understanding of the environment and more efficient learning trajectories. The implications are significant for V2X vehicular communication problems where rapid learning is crucial due to the dynamic nature of vehicular environments and the critical need for safety.

\subsection{Robustness}

When we deal with real world scenarios, one should consider that the environment varies across time due to its non-stationary behavior. Thus, in order to evaluate the robustness of the DRL algorithms tested above to the variation of the environment, we propose to analyze the generalization affinity of the policies by picking up the two bests learned policies of each algorithm and tested  their performance in similar configuration with scenario $2$ . Table~\ref{table:Generalisation}  summarized the performance in terms of reliability obtained with the unseen scenario. As we can observe, for the 1$^{\text{st}}$ best models of the benchmark, there is an important gap of performance between the training dataset and new data. However, the 2$^{\text{nd}}$ best model approaches better the previous performance on the training environment. This phenomena is caused by overfitting, which happen when the model is overtrained on the training environment. Traditionaly, for machine learning model, the overfitting is avoided by adding a regularization technique. In the context of RL, there are 4 different regularization technique that we mention in the performance analysis section.

\section{Conclusion and future works}
\label{conclusion}


In this article, we offer a benchmark of DRL algorithms adapted to the vertical handover problem in the V2X context. The role of the agent is to select the optimal access points for each transmission. We select a couple of DRL agents, train and fine tune them with a simple grid search. The best model was tested and evaluated in term of reliability (most of them achieve the level of quality that is required by 3GPP standard in order to support advanced V2X applications). The strategy learned from each algorithm has been visually represented, enabling in-depth analysis of the algorithm's effectiveness and stability.  The study showed that the PPO is the best model for reliability, stability, robustness, complexity, and number of switches followed by SAC, Rainbow, TRPO. However, we discuss that we can further decrease this number by introducing the hysteresis filter. Lastly, to give an idea of real-world performance, we provide a test of robustness by changing the environment. In future work, we plan to extend this study by adding additional access points such as C-V2X, test other scenarios (highway, intersection) and have more accurate representations of the vehicle environment by introducing more modalities.

\bibliography{aaai24}

\end{document}